\pdfoutput=1

\documentclass[11pt]{article}

\usepackage[]{acl}

\usepackage{times}
\usepackage{latexsym}

\usepackage[T1]{fontenc}

\usepackage[utf8]{inputenc}

\usepackage{microtype}

\usepackage{amsmath,amsfonts,amssymb}
\usepackage{amsthm}
\usepackage{graphicx}
\usepackage{bm}
\usepackage{cases}
\usepackage{multirow}
\usepackage{subfigure}
\usepackage{booktabs}
\usepackage{mathtools}
\usepackage{color}
\usepackage{stmaryrd}
\usepackage{pifont}
\usepackage{changes}

\usepackage{inconsolata}

\title{Mitigating  Heterogeneity among Factor Tensors via Lie Group Manifolds for Tensor Decomposition Based Temporal Knowledge Graph Embedding}

\author{Jiang Li$^{1,2,3}$ ,  Xiangdong Su$^{1,2,3}$ \thanks{\ \ Corresponding Author}, Guanglai Gao$^{1,2,3}$ \\
$^1$ College of Computer Science, Inner Mongolia University, China \\ $^2$ National \& Local Joint Engineering Research Center of Intelligent Information \\ Processing Technology for Mongolian, China\\ $^3$ Inner Mongolia Key Laboratory of Multilingual Artiffcial Intelligence Technology, China \\ \texttt{lijiangimu@gmail.com, cssxd@imu.edu.cn}}

\begin{document}
\maketitle
\begin{abstract}

Recent studies have highlighted the effectiveness of tensor decomposition methods in the Temporal Knowledge Graphs Embedding (TKGE) task. However, we found that inherent heterogeneity among factor tensors in tensor decomposition significantly hinders the tensor fusion process and further limits the performance of link prediction. To overcome this limitation, we introduce a novel method that maps factor tensors onto a unified smooth Lie group manifold to make the distribution of factor tensors approximating homogeneous in tensor decomposition. We provide the theoretical proof of our motivation that homogeneous tensors are more effective than heterogeneous tensors in tensor fusion and approximating the target for tensor decomposition based TKGE methods. The proposed method can be directly integrated into existing tensor decomposition based TKGE methods without introducing extra parameters. Extensive experiments demonstrate the effectiveness of our method in mitigating the heterogeneity and in enhancing the tensor decomposition based TKGE models\footnote{Our code is available at \url{https://github.com/dellixx/tkbc-lie}}.

\end{abstract}

\section{Introduction}

Knowledge graphs (KGs) are data structures that encapsulate knowledge triples of real-world entities and their interrelationships, and are widely used to improve information retrieval~\cite{1}, reasoning~\cite{2}, Q\&A~\cite{3}, etc. Temporal knowledge graphs (TKGs) extend this paradigm by introducing timestamps into knowledge triplets to reflect the validity of facts over time and provide a deeper understanding and analysis of dynamic changes in the facts. 
Due to the data incompleteness in both KGs and TKGs, researchers propose many KG embedding (KGE) and TKG embedding (TKGE) methods to predict the missing facts, thereby enhancing the richness and accuracy of the KGs and TKGs. This work mainly focuses on TKGE.

\begin{figure}[t]
    \centering
    \includegraphics[width=0.9\linewidth]{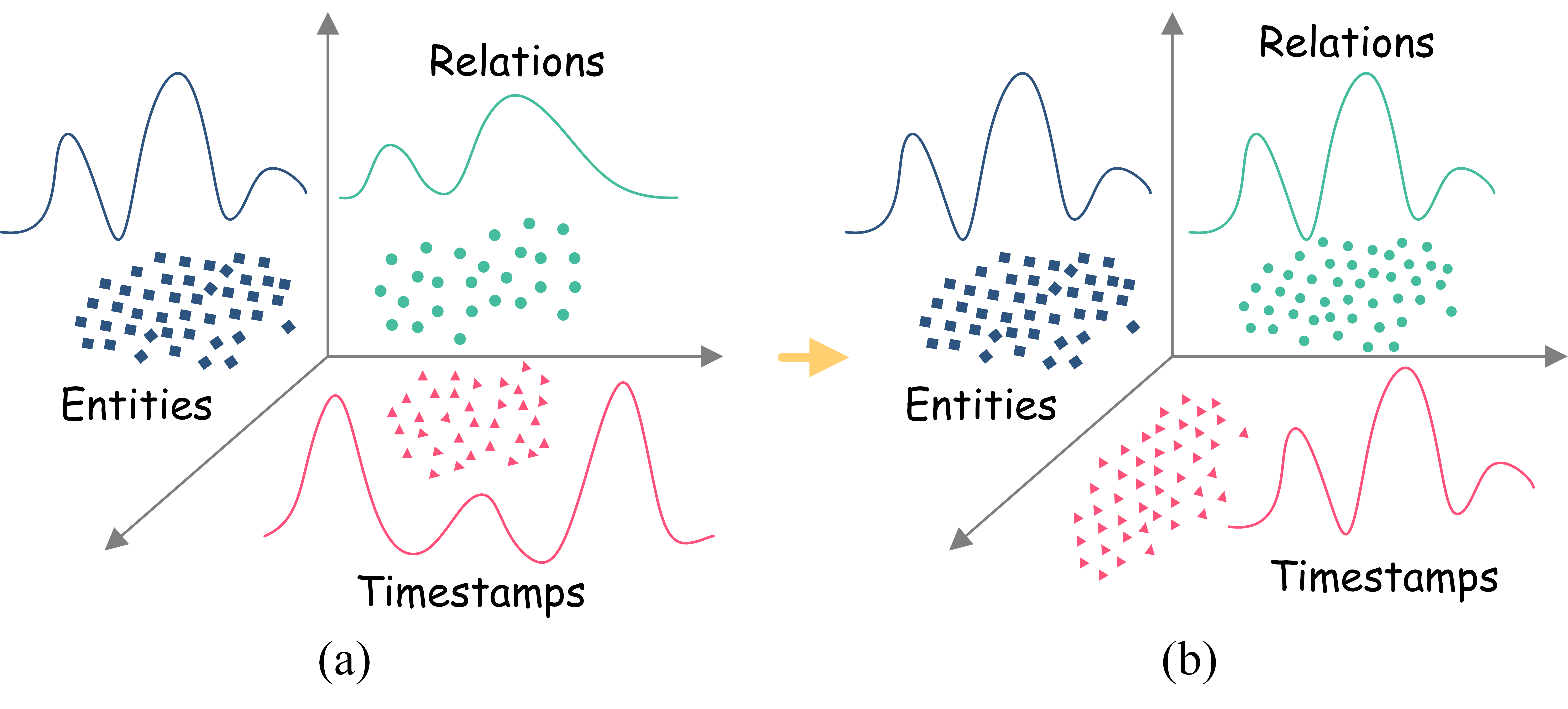}
    \caption{(a) illustrates the heterogeneity in the distribution of entities, relations and timestamps within TKGs, as evidenced by the differing distribution curves. (b) illustrates the homogeneous distribution curves of entities, relations and timestamps when using our method to mitigate the heterogeneity among these three elements.}
    \label{fig:start}
\end{figure}

As the interest in TKG grows, researchers proposed many TKGE methods and greatly promoted the development of TKG.
Concerning the success of tensor decomposition in KGE~\cite{nickel2016review, 2016-complex, 2018-n3}, recent works~\cite{2019-tnt,2021-telm,li-etal-2023-teast} further extended tensor decomposition into TKGE and obtained very excellent performance. 
These works demonstrated that tensor decomposition can guarantee full expressiveness under specific embedding dimensionality bounds in TKG, thus enhancing the link prediction.

However, existing TKGE methods based on tensor decomposition suffer from inherent heterogeneity among factor tensors. Recent research~\cite{wu2020temp,li2021learning} also highlights the intrinsic heterogeneity in TKGs, specifically in terms of entity and temporal heterogeneity. According to our analysis, the heterogeneity of entity, relation and timestamp originates from their semantic roles within the knowledge graphs. That is, the entities represent the static components of the graph, the relations delineate the interactions among the entities, and the timestamp characterizes the temporal aspects of these interactions, specifying when they occur and their duration. This heterogeneity leads to the learned factor tensor expliciting different distributions in TKGE, as shown in Figure~\ref{fig:start}(a). This further limits the tensor fusion in TKGE models and lowers the link prediction accuracy. More discussion about heterogeneity can be found in \textbf{Appendix~\ref{sec:ddh}}.

Therefore, it is necessary to address the heterogeneity for tensor decomposition-based TKGE methods to enhance link prediction. To this target, we propose to map the factor tensors onto a unified smooth Lie group manifold to make the distribution of factor tensors approximating homogeneous in tensor decomposition, as shown in Figure~\ref{fig:start}(b). 
Since the manifold in Lie group looks the same at every point and all tangent spaces at any point are alike~\cite{sola2018micro}, the factor tensors mapped by the Lie group have a smooth and unified distribution, which mitigates the heterogeneity among the factor tensors. We provide the theoretical proof of our motivation that homogeneous factor tensors are more effective in approximating the target compared to heterogeneous factor tensors in TKGE models in Sec.~\ref{sec:4.1}. We integrate the proposed method into several existing tensor decomposition based TKGE models and conduct extensive experiments to evaluate its effectiveness. The experimental results present the heterogeneity among factor tensors in TKGE methods and illustrate that the proposed method brings significant performance improvement. This confirms the effectiveness of our method in alleviating the heterogeneity. Our contributions are summarized as follows:
\begin{itemize}
    \item To the best of our knowledge, we are the first to investigate the negative effect of the heterogeneity among the factor tensors for tensor decomposition based TKGE models and propose to enhance these models by diminishing the heterogeneity via Lie group manifold.
    \item We provide the theoretical proof of our motivation that homogeneous factor tensors are more effective than heterogeneous factor tensors in approximating the target in TKGE.
    \item  Our proposed method can be directly integrated into the tensor decomposition based TKGE models without introducing any additional parameters, and extensive experiments on several TKGE models demonstrate its effectiveness and generalization.
\end{itemize}

\section{Related Work}

\subsection{Static Knowledge Graph Embedding}
Drawing inspiration from the concept of  translation invariance featured in word2vec~\cite{2013word2vec}, TransE assesses the relations between entities and their links by calculating the distance from $\bm{e_s}+\bm{e_{\hat{r}}}$ to $\bm{e_o}$ using standard $l_1$ or $l_2$ norms, where  $\bm{e_s}$ and $\bm{e_o}$ are the vectors that represent the starting and ending entities, and $\bm{e_{\hat{r}}}$ represents the linking relation.   Following TransE, TransH~\cite{2014-transh}, TransR~\cite{2015-transr}, and TransD~\cite{2015-transd}  introduce various mapping ways and thus refine these embeddings for better KGE representation. ComplEx~\cite{2016-complex}  employs 3-th order tensor decomposition to capture the interactions within KGs. TorusE~\cite{ebisu2018toruse} utilizes a torus (a donut-shaped manifold) for its embeddings. TorusE introduces a torus, which is a compact Abelian Lie group, and defines distance functions on the torus. The torus can be considered as a collection of multiple Lie groups. Instead, we map the factor tensors to the Lie group space, thus mitigating the distributional heterogeneity among them. \\

\subsection{Temporal Knowledge Graph Embedding}
In TKGE models, the temporal information is added, and the scoring function is calculated for the quadruples to assess their reasonableness. Therefore, most TKGE models use existing KGE models as a foundation. TTransE~\cite{2018-ttranse} extends TransE and encodes time stamps $\tau$ as translations same as relations. Hence, the score function of TTransE is denoted as $\phi (s,\hat{r},o,\tau) = ||\bm{s}+\bm{{\hat{r}}}+\bm{\tau}-\bm{o}||_p$.  Furthermore, TA-TransE~\cite{garcia2018learning} encode timestamps based on TransE. RotateQVS~\cite{2022-rotateqvs} uses quaternion embeddings to represent both entities and relations. Recently, BoxTE~\cite{messner2022temporal} models the TKGE based on a box embedding model BoxE~\cite{abboud2020boxe}. TCompoundE~\cite{2024-TCompoundE} employs relation-specific and time-specific compound geometric operations to enhance the modeling of temporal dynamics and relational patterns. \\


\subsection{Tensor Decomposition Based Temporal Knowledge Graph Embedding}

TComplEx~\cite{2019-tnt} and TNTComplEx~\cite{2019-tnt} expand upon the ComplEx model by executing a fourth-order tensor decomposition in temporal knowledge graphs (TKGs). This method offers a more nuanced understanding of the temporal dimensions in knowledge graphs. TeLM~\cite{2021-telm}  utilizes the asymmetric geometric product, a method that allows for a more sophisticated and expressive representation of temporal relationships and entities. TeAST~\cite{li-etal-2023-teast}  maps relations onto Archimedean spiral timelines, and ensures that relations occurring simultaneously are placed on the same timeline, with all relations evolving over time. The above works are all based on tensor decomposition to optimize the TKGE representation. In this paper, we focus on exploring the problem of the heterogeneity of factor tensors in tensor decomposition based TKGE models.

\section{Background and Notation}

\subsection{TKGE Task}
Given a TKG, let $\mathcal{E}$ denote the set of entities, $\mathcal{R}$ denote the set of relations, and $\mathcal{T}$ denote the set of timestamps. A TKG can be defined as a collection of quadruples $(s,\hat{r},o,\tau)$, where $s\in \mathcal{E}$, $\hat{r}\in\mathcal{R}$, $o \in\mathcal{E}$ and $\tau\in\mathcal{T}$ denote the subject entity, relation, object entity and timestamp, respectively. The TKGE task aims to accurately learn embedded representations of entities, relations and timestamps to facilitate predictions of missing entities in TKGs. Specifically, it involves predicting the object entity \(o\) given a tuple $(s, \hat{r}, ?, \tau)$, or conversely, predicting the subject entity \(s\) for a tuple $(?, \hat{r}, o, \tau)$, thereby capturing the dynamic nature of relationships over time. In this paper, we denote the relation in TKG as $\hat{r}$ and the rank in mathematics as $r$.

\subsection{Tensor Decomposition for TKGE}
In the existing tensor decomposition based TKGE methods, each relation quadruple is represented by a $\{0,1\}$-valued 4-th order tensor \(\mathcal{Y} \in \{0,1\}\)~\cite{2019-tnt}. This representation allows each element \(\mathcal{Y}_{s,\hat{r},o,\tau}=1\) to indicate that at a specific time $\tau$, there is a relationship  \(\hat{r}\) between entities \(s\) and \(o\). 
In link prediction, tensor decomposition algorithms learn to infer a predicted tensor $\mathcal{X}$ that approximates the ground truth $\mathcal{Y}$, as

\begin{equation}\label{eq1}
\mathcal{Y}\sim\mathcal{X} = \sum_{r=1}^{R}  \mathbf{u}_r \otimes \mathbf{v}_r \otimes   \mathbf{w}_r \otimes \mathbf{t}_r,
\end{equation}
where rank $r\in\{1,...,R\}$, $\otimes$ denotes the tensor product. $\mathbf{u}_r$,  $\mathbf{v}_r$, $\mathbf{w}_r$ and $\mathbf{t}_r$ denote the subject entity, relation, object entity and timestamp factor tensors. Tensor decomposition based TKGE methods aim to optimize the factor tensors to make $\mathcal{X}$ as close as possible to the tensor $\mathcal{Y}$ and thus achieve more accurate link prediction.

\section{Methodology}
This paper employs Lie group manifold to diminish the heterogeneity of factor tensors in tensor decomposition based TKGE models and thus improves the performance of these models. \textbf{In Sec.~\ref{sec:4.1}}, we provide the theoretical proof of our motivation that homogeneous tensors are more effective than heterogeneous tensors in approximating the target for tensor decomposition based TKGE methods. \textbf{In Sec.~\ref{sec:4.2}}, we explain why Lie groups can mitigate the heterogeneity among factor tensors, and describe how to map the factor tensors to the Lie group space.
\textbf{In Sec.~\ref{sec:4.3}}, we introduce a Logarithmic Mapping `${log}(f(\cdot))$' operation and alleviate the heterogeneity among factors by minimizing the difference between the original factor tensors and the mapped factor tensors in the Lie group through the N3 regularization in the loss function.

\subsection{Theoretical Analysis of Homogeneous \textit{vs}. Heterogeneous Factor Tensors}\label{sec:4.1}

\textbf{Proposition 1.} \emph{Homogeneous factor tensors ($\mathbf{u}_r$,  $\mathbf{v}_r$, $\mathbf{w}_r$, $\mathbf{t}_r$) with a low rank can effectively approximated $\mathcal{Y}$ while heterogeneous factor tensors ($\mathbf{u}_r$, $\mathbf{v}_r$, $\mathbf{w}_r$,  $\mathbf{t}_r$) require a higher rank to approximate $\mathcal{Y}$ in TKGE.}\\\\
\emph{\textbf{Proof.}} Given a 4th-order tensor decomposition in TKGE

\begin{equation}
\mathcal{Y} \approx \sum_{r=1}^{R}  \mathbf{u}_r \otimes \mathbf{v}_r \otimes   \mathbf{w}_r \otimes \mathbf{t}_r,
\end{equation}
where \( R \) is the rank of the decomposition, and $(\mathbf{u}_r,\mathbf{v}_r,\mathbf{w}_r,\mathbf{t}_r) \in \mathbb{R}$ are factor tensors.

If the factor tensors ($\mathbf{u}_r$, $\mathbf{v}_r$, $\mathbf{w}_r$, $\mathbf{t}_r$) are homogenous, 
there are a common set of basis vectors \( \mathbf{B} = \{ \mathbf{b}_1, \mathbf{b}_2, \ldots, \mathbf{b}_m \} \) between these factor tensors. 
Based on this homogeneity, the column vectors of each factor matrix can be expressed as linear combinations of the basis vectors in \( \mathbf{B} \).

\begin{equation}
\begin{split}
\mathbf{u}_r = \mathbf{B} \cdot \boldsymbol{\alpha}_r,\quad 
\mathbf{v}_r = \mathbf{B} \cdot \boldsymbol{\beta}_r,  \\
\mathbf{w}_r = \mathbf{B} \cdot \boldsymbol{\gamma}_r, \quad 
\mathbf{t}_r = \mathbf{B} \cdot \boldsymbol{\delta}_r,
\end{split}
\end{equation}
where \( r = 1, \ldots, R \), and \( \boldsymbol{\alpha}_r \), \( \boldsymbol{\beta}_r \), \( \boldsymbol{\gamma}_r \), and \( \boldsymbol{\delta}_r \) are the coefficient vectors for the \( r \)-th component in their respective factor matrices. 
Consequently, we can get 

\begin{equation}
\scalebox{0.85}{$
\mathcal{Y} \approx \sum_{r=1}^{R}   (\mathbf{B} \cdot \boldsymbol{\alpha}_r) \otimes (\mathbf{B} \cdot \boldsymbol{\beta}_r) \otimes (\mathbf{B} \cdot \boldsymbol{\gamma}_r) \otimes (\mathbf{B} \cdot \boldsymbol{\delta}_r).$}
\end{equation}

 Given the homogeneity among factor tensors, we can further obtain the representation as follows

\begin{equation}
\scalebox{0.85}{$
\mathcal{Y} \approx \sum_{j=1}^{m}  \mathbf{b}_j \otimes \mathbf{b}_j \otimes \mathbf{b}_j \otimes \mathbf{b}_j \cdot \sum_{r=1}^{R} \alpha_{jr} \beta_{jr} \gamma_{jr} \delta_{jr},$}
\end{equation}
where \( \alpha_{jr} \), \( \beta_{jr} \), \( \gamma_{jr} \), and \( \delta_{jr} \) are the scalar coefficients corresponding to the projection of the \( r \)-th component onto the \( j \)-th basis vector in their respective factor matrices. The set of basis vectors \(\{\mathbf{b}_1, \mathbf{b}_2, \ldots, \mathbf{b}_m\}\) are orthogonal to each other, ensuring that each dimension represented by these vectors is independent.  \( \lambda_j = \sum_{r=1}^{R} \alpha_{jr} \beta_{jr} \gamma_{jr} \delta_{jr} \) can be considered as scaling constants. Thus, we can get a reduced-rank tensor \( \mathcal{Y}' \) as

\begin{equation}
\mathcal{Y} \approx \mathcal{Y}' = \sum_{j=1}^{m} \lambda_j \mathbf{b}_j \otimes \mathbf{b}_j \otimes \mathbf{b}_j \otimes \mathbf{b}_j,
\end{equation}
which captures the essence of the homogeneity within the factor matrices. The rank \( m \) of \( \mathcal{Y}' \) is less than  the original rank $R$ ($m<R$).
Thus, homogeneous factor tensors ($\mathbf{u}_r$,  $\mathbf{v}_r$, $\mathbf{w}_r$, $\mathbf{t}_r$) with a low rank can effectively approximate the target quadruple tensor \( \mathcal{Y} \).

In contrast, if the factor tensors ($\mathbf{u}_r$, $\mathbf{v}_r$, $\mathbf{w}_r$, $\mathbf{t}_r$) are highly heterogeneous, they exhibit distinct semantics and distributions characteristics. This implies that these factor tensors cannot be effectively approximated by a simple, small number of rank-1 tensors. Each rank-1 tensor can be viewed as a representation of a specific pattern or feature. 
For these heterogeneous tensors, the patterns they capture, or the semantics they represent within the data necessitate a larger number of rank-1 tensors to capture their diverse characteristics individually. Hence, the heterogeneous factor tensors ($\mathbf{u}_r$, $\mathbf{v}_r$, $\mathbf{w}_r$,  $\mathbf{t}_r$) require a higher rank or even  full rank to approximate $\mathcal{Y}$ in TKGE. Since higher rank means more parameters to estimate and more computation, homogeneous factor tensors are more effective than heterogeneous factor tensors in approximating the target for tensor decomposition based TKGE methods.

\begin{figure}[t]
    \centering
    \includegraphics[width=0.81\linewidth]{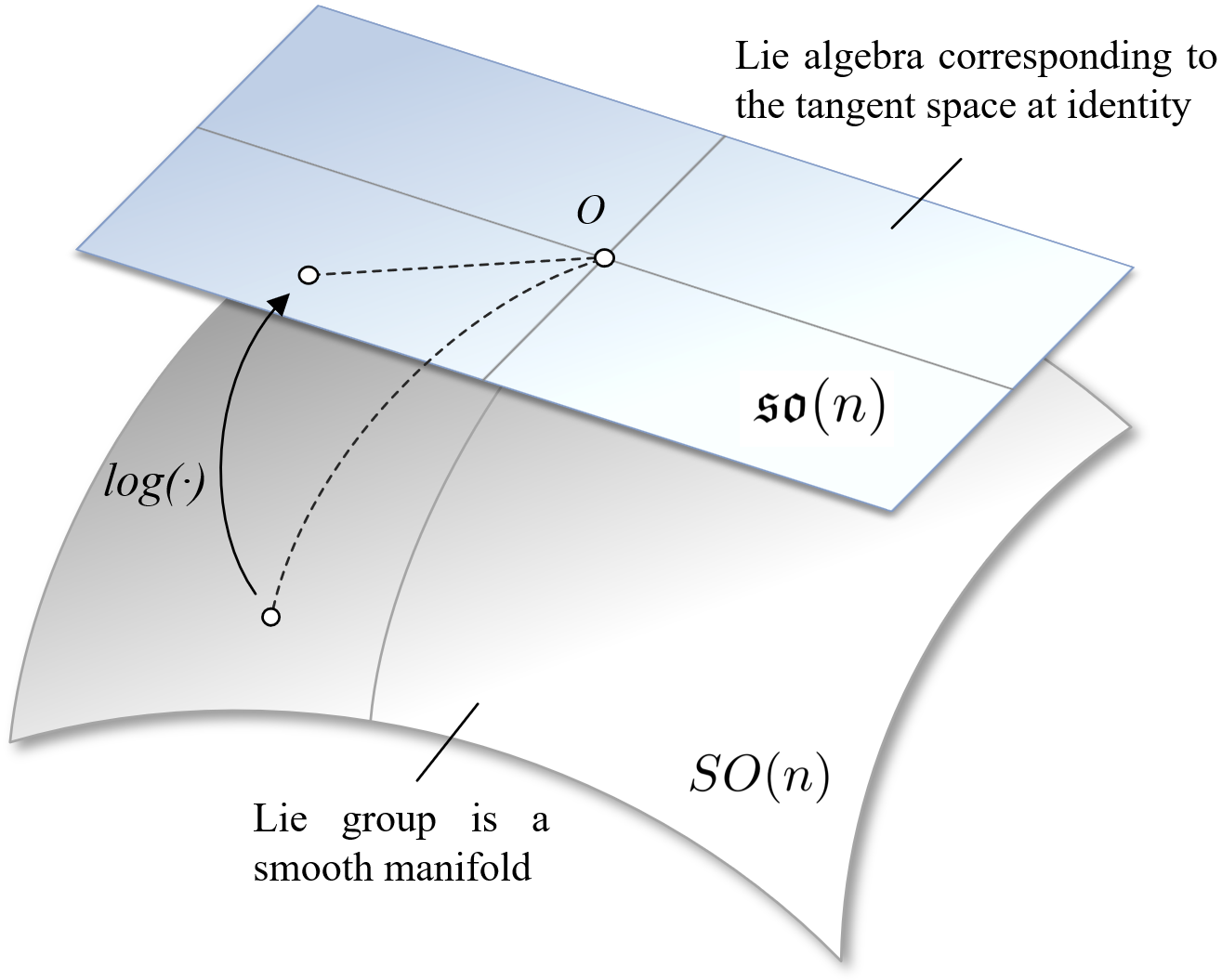}
    \caption{An illustration of the relation between the Lie group and the Lie algebra. The Lie algebra $\mathfrak{so}(n)$ is the tangent space to the Lie group’s manifold $S(n)$.}
    \label{fig:enter-label}
\end{figure}

\subsection{Mitigating Heterogeneity via Lie Group}\label{sec:4.2}

In TKGE, the underlying reason why highly heterogeneous factor tensors require a higher rank to approximate the target vector is that the heterogeneity among factor tensors can limit the fusion process of subsequent computations, which can be analogous to the multimodal fusion process~\cite{chen2020hgmf}. Therefore, it is crucial to mitigate the heterogeneity among factor tensors to approximate the target tensor more efficiently. Our motivations for choosing Lie groups to mitigate heterogeneity in TKGE are as follows.

Firstly, Lie groups are adept at maintaining structural integrity and handling data's dynamic nature over time, making them a suitable choice for TKGE. The application of Lie groups in fields like robotics~\cite{sola2018micro}, machine learning~\cite{sommer2020efficient,lee2022hara}, and computer vision~\cite{teed2021tangent} underscores its capability to model complex geometric transformations effectively.  Secondly, as shown in Figure~\ref{fig:enter-label}, Lie group is a mathematical structure that simultaneously satisfies the axioms of a group and the properties of a smooth manifold. It is like a curved, smooth hyper-surface, with no edges or spikes, embedded in a space of higher dimension.  The smoothness of the manifold implies the existence of a unique tangent space at each point. In a Lie group, the manifold looks the same at every point, and therefore all tangent spaces at any point are alike. Thus, the factor tensors mapped by the Lie group have a smooth and unified distribution, which further mitigates the heterogeneity among the factor tensors.

In this study, we map the factor tensors to the same Lie group space and make the factor tensors have a unified distribution to mitigate the heterogeneity among them. To facilitate the description of our method, we employ  factor tensor of rank 4 in the subsequent discussions.
Given a factor tensor $\bm{e}$ of rank 4 and map it to the Lie group $SO(2)$ space, we get

\begin{equation}
    f(\cdot): \mathbb{R}^n \to SO(2); \quad \bm{e} \mapsto \mathbf{R}
\end{equation}
where $f(\cdot)$ denots Lie group mapping operation. $\mathbf{R}_{\bm{e}} $ is a rotation matrix in $SO(2)$, denoted as

\begin{equation}
\scalebox{0.95}{$
\mathbf{R}_{\bm{e}} = \begin{pmatrix} \cos \bm{e} & -\sin \bm{e} \\ \sin \bm{e} & \cos \bm{e}  \end{pmatrix}.$}
\end{equation}

Accordingly, given a quadruple ($s$, ${\hat{r}}$, $o$, $\tau$) in TKG, its corresponding factor tensors  are ($\mathbf{u}$,  $\mathbf{v}$, $\mathbf{w}$, $\mathbf{t}$) in the tensor decomposition based TKGE models. We map these four factor tensors  onto Lie group space, and we get the rotation matrices

\begin{equation}
\scalebox{0.9}{$
\begin{split}
    \mathbf{R_u} = \begin{pmatrix} \cos \mathbf{u} & -\sin \mathbf{u} \\ \sin \mathbf{u} & \cos \mathbf{u} \end{pmatrix}, 
    \mathbf{R_v} = \begin{pmatrix} \cos \mathbf{v} & -\sin \mathbf{v} \\ \sin \mathbf{v} & \cos \mathbf{v} \end{pmatrix},\\
    \mathbf{R_w} = \begin{pmatrix} \cos \mathbf{w} & -\sin \mathbf{w} \\ \sin \mathbf{w} & \cos \mathbf{w} \end{pmatrix}, 
    \mathbf{R_t} = \begin{pmatrix} \cos \mathbf{t} & -\sin \mathbf{t} \\ \sin \mathbf{t} & \cos \mathbf{t} \end{pmatrix}.  
\end{split}$}
\end{equation}

When generalized to \( n \) rank,  \( SO(\sqrt{n}) \) can be denoted as a $Givens$ rotation  matrix. In an \( n \)-dimensional space, a Givens rotation is performed by fixing \( \sqrt{n}-2 \) dimensions and applying a rotation transformation within the plane formed by the remaining two dimensions. This effectively isolates the rotation to a specific two-dimensional subspace. The Givens rotation matrix $G(i, j, \bm{e})$ for rotating the \( i \)-th and \( j \)-th coordinates in \( \sqrt{n} \)-dimensional space is given by:

\begin{equation}
\scalebox{0.71}{
$G(i, j, \bm{e}) = 
\begin{bmatrix}
1 & \cdots & 0 & \cdots & 0 & \cdots & 0 \\
\vdots & \ddots & \vdots & & \vdots & & \vdots \\
0 & \cdots & \cos(\bm{e}) & \cdots & -\sin(\bm{e}) & \cdots & 0 \\
\vdots & & \vdots & \ddots & \vdots & & \vdots \\
0 & \cdots & \sin(\bm{e}) & \cdots & \cos(\bm{e}) & \cdots & 0 \\
\vdots & & \vdots & & \vdots & \ddots & \vdots \\
0 & \cdots & 0 & \cdots & 0 & \cdots & 1
\end{bmatrix}$
},
\end{equation}
where the rotation occurs in the plane spanned by the \( i \)-th and \( j \)-th basis vectors. The matrix is an identity matrix except for the four elements \( G_{ii} \), \( G_{ij} \), \( G_{ji} \), and \( G_{jj} \), which form the $2\times2$ rotation block within the larger matrix. Hence,  we focus on the $2\times2$ base rotation matrices in the code implementation.

\subsection{Logarithmic Mapping from $SO(n)$ to $\mathfrak{so}(n)$ } \label{sec:4.3}

Our training goal is to diminish the heterogeneity among the factor tensors in TKGE model training and thus improve the link prediction performance. 
Due to the Lie group $SO(n)$ residing on a non-Euclidean manifold, we introduce a variant of Logarithmic Mapping operation~\cite{Huang_2017_CVPR} (denoted as $log(\cdot)$) on the Lie group space and converts the rotation matrices into the usual skew-symmetric matrices which are situated in the Euclidean space, as 

\begin{equation}
    {log}(\cdot): SO(n) \to \mathfrak{so}(n),\quad \mathbf{R} \mapsto log(\mathbf{R}).
\end{equation}

The logarithm mapping $log(\mathbf{R})$ is

\begin{equation}
\scalebox{0.95}{$
{log}(\mathbf{R}) = 
\begin{cases} 
0, & \text{if } \quad \theta(\mathbf{R}) = 0, \\
\frac{\theta(\mathbf{R})}{2\sin(\theta(\mathbf{R}))}(\mathbf{R} - \mathbf{R}^T), & \text{otherwise},
\end{cases} $}
\end{equation}
where $\theta(\mathbf{R})$ is the angle of $\mathbf{R}$, as
\begin{equation}\label{eq15}
\theta(\mathbf{R}) = \arccos\left(\frac{trace(\mathbf{R}) - 1}{2}\right).
\end{equation}

Here, $trace(\cdot)$ is a square matrix is the sum of its diagonal elements.

The mathematical derivation of logarithmic mappings references this work~\cite{sola2018micro}. After logarithmic mapping, we get $log(f(\mathbf{u}_r))$, $log(f(\mathbf{v}_r))$, $log(f(\mathbf{w}_r))$, $log(f(\mathbf{t}_r))$. Then, we calculate the differences between the original tensors and their corresponding mapped tensors on the Lie group  

\begin{equation}
\scalebox{0.99}{
$\begin{split}
\mathbf{u}_r' &=  \mathbf{u}_r - log(f(\mathbf{u}_r)), \\ 
\mathbf{v}_r' &= \mathbf{v}_r - log(f(\mathbf{v}_r)),  \\
\mathbf{w}_r' &= \mathbf{w}_r - log(f(\mathbf{w}_r)),  \\
\mathbf{t}_r' &= \mathbf{t}_r - log(f(\mathbf{t}_r)).
\end{split}$}
\end{equation}

Finally, we perform the standard tensor decomposition with $\mathbf{u}_r'$, $\mathbf{v}_r'$, $\mathbf{w}_r'$, $\mathbf{t}_r'$, as

\begin{equation}
\mathcal{Y}\sim\mathcal{X} = \sum_{r=1}^{R} \lambda_r \mathbf{u}_r' \otimes \mathbf{v}_r' \otimes   \mathbf{w}_r' \otimes \mathbf{t}_r'.
\end{equation}

Following previous works~\cite{2018-n3,2019-tnt,2021-telm,li-etal-2023-teast}, we use the full multiclass log-softmax loss function and N3 regularization to optimize the factor tensors, which are defined as follows:

\begin{equation}\label{eq8}
\scalebox{0.99}{$
	\begin{split}
		\mathcal{L } = -\log(\frac{\exp(\phi (s,\hat{r},o,\tau))}{\sum_{s'\in\mathcal{E}} \exp(\phi (s',\hat{r},o,\tau)) }) \\- \log(\frac{\exp(\phi (o,\hat{r}^{-1},s,\tau))}{\sum_{o'\in\mathcal{E}} \exp(\phi (o',\hat{r}^{-1},s,\tau)) })\\ +\lambda_\mu \sum_{i=1}^R(\Vert \mathbf{u}_r' \Vert_3^3 + \Vert \mathbf{v}_r' \Vert_3^3 + \Vert \mathbf{w}_r' \Vert_3^3 + \Vert \mathbf{t}_{r}' \Vert_3^3 ),
	\end{split}$}
\end{equation}
where $\lambda_\mu$ denotes N3 regularization weight, $\hat{r}^{-1}$ is the inverse relation. We use $\Vert \mathbf{t}_{r}' \Vert_3^3$
 to represent the temporal regularizer for simplicity, which is computed in N3 regularization way.

By using the N3 regularization, we minimize the $\mathbf{u}_r'$, $\mathbf{v}_r'$, $\mathbf{w}_r'$, $\mathbf{t}_r'$. That is, we drive the $\mathbf{u}_r$, $\mathbf{v}_r$, $\mathbf{w}_r$, $\mathbf{t}_r$ to be homogeneous in Euclidean space, since the $log(f(\mathbf{u}_r))$, $log(f(\mathbf{v}_r))$, $log(f(\mathbf{w}_r))$, $log(f(\mathbf{t}_r))$ are tend to be homogeneous in Lie group space. Therefore, the proposed method can mitigate the heterogeneity among factor tensors in tensor decomposition based TKGE methods.

\section{Experiments}
\subsection{Datasets}

To evaluate the effectiveness of the proposed method, we evaluate our method on two popular TKGE benchmark datasets. ICEWS14 and ICEWS05-15~\cite{garcia2018learning} are both extracted from the \emph{Integrated Crisis Early Warning System (ICEWS)} dataset~\cite{sbdata}, which consists of temporal sociopolitical facts starting from 1995. ICEWS14 consists of sociopolitical events in 2014 and ICEWS05-15 involves events occurring from 2005 to 2015. ICEWS14 is a fine temporal granularity dataset, while ICEWS05-15 has a wider temporal granularity relative to ICEWS14.  See \textbf{Appendix~\ref{sec:appendix}} for summary statistics of the dataset and more discussion of the dataset.

\subsection{Evaluation Protocol}
In this research, we follow the previous works~\cite{2019-tnt,2021-telm,li-etal-2023-teast} to evaluate our method. Specifically, to evaluate the quality of the ranking for each test quadruples, we calculate all possible substitutions for the subject and object entities, denoted as $(s',\hat{r},o,\tau)$ and $(s,\hat{r},o',\tau)$, where $s'$, $o'\in \mathcal{E}$.  After that, we sort the score of candidate quadruples under the time-wise filtered settings~\cite{2019-tnt, 2021-telm,li-etal-2023-teast}. The performance is evaluated using standard evaluation metrics, including Mean Reciprocal Rank (MRR) and Hits@$n$. The Hits@$n$ metric measures the percentage of correct entities in the top $n$ predictions. Higher values of MRR and Hits@$n$ indicate better performance. Hits ratio with cut-off values $n = 1, 3, 10$. In this paper, we utilize H@$n$ to denote Hits@$n$ for convenience.

\subsection{Experimental Setup}
We implement our method based on the existing training framework\footnote{\url{https://github.com/facebookresearch/tkbc}}. All experiments are trained on a single NVIDIA Tesla A100. The hyperparameters used in the experiment are consistent with the optimal hyperparameters of the original paper report. The best models are selected by early stopping (threshold of 10) on the validation datasets. The max epoch is 200. We report the average results on the test set for five runs. To ensure a fair validation of the effectiveness of our method, we employ the same hyperparameter configuration in both the before and after comparison experiments.

According to Lie group mapping described in Sec.~\ref{sec:4.2}, it is essential to ensure that the rank $r$ of the matrix can be satisfied by the square root of $(2 \times  r)$ is an integer in our method. This requirement arises from the implementation of the matrix logarithm map for TcomplEx, TNTcomplEx and TeAST. The rank of TeLM needs to be satisfied by the square root of $(4\times r)$ is an integer.

\section{Results and  Analysis}
  
\subsection{Main Results}

\linespread{1.2}
\begin{table*}[!htb]
\centering
\scalebox{0.83}{
\begin{tabular}{lcccccccccccc}
    \specialrule{1.2pt}{0pt}{0pt}
           & \multicolumn{6}{c}{\textbf{ICEWS14}}                  & \multicolumn{6}{c}{\textbf{ICEWS05-15}}               \\ \cmidrule(lr){2-7} \cmidrule(lr){8-13} 
           & \emph{rank} & \emph{Para.} & MRR   & H@1   & H@3   & H@10  & \emph{rank} & \emph{Para.} & MRR   & H@1   & H@3   & H@10  \\ \hline
    \multicolumn{13}{l}{\emph{\textbf{Tensor Decomposition Based TKGE Models}}} \\
    \hline
TComplEx$^\heartsuit$   & 128  & 2.04M & 55.3  & 46.3  & 60.7  & 71.5  & 128  & 3.84M & 58  & 49  & 64  & 76  \\
TNTComplEx$^\heartsuit$ & 128  & 2.15M & 55.7  & 46.3  & 61.5  & 73.0  & 128  & 3.97M & 60  & 50  & 65  & 78  \\
TeLM       & 121  & 3.85M & 50.6 & 42.1 & 55.0 & 67.1 & 121  & 7.26M & 56.8 & 48.7 & 61.1 & 72.0 \\
TeAST      & 128  & 2.13M & 53.4 & 43.9 & 58.8 & 70.9 & 128  & 4.87M & 48.8 & 38.4 & 54.7 & 68.3 \\ \hline

    \multicolumn{13}{l}{\emph{\textbf{Tensor Decomposition Based TKGE Models$+{log}(f(\cdot))$}}} \\
    \hline
\multirow{2}{*}{TComplEx}   & \multirow{2}{*}{128}  & \multirow{2}{*}{2.04M} & \textbf{56.2} & \textbf{46.8} & \textbf{61.6} & \textbf{73.5} & \multirow{2}{*}{128}  & \multirow{2}{*}{3.84M} & \textbf{59.6} & \textbf{50.2} & \textbf{65.3} & \textbf{77.2} \\
&          &       &   \textcolor{red}{\small{( +0.9)}}    &   \textcolor{red}{\small{( +0.5)}}    & \textcolor{red}{\small{( + 0.9)}} & \textcolor{red}{\small{( +0.9)}} &          &       &   \textcolor{red}{\small{( +1.6)}}    &   \textcolor{red}{\small{( +0.3)}}    & \textcolor{red}{\small{( +1.3)}} & \textcolor{red}{\small{( +1.2)}} \\ \hline
\multirow{2}{*}{TNTComplEx} & \multirow{2}{*}{128}  & \multirow{2}{*}{2.15M} & \textbf{56.3} & \textbf{46.7} & \textbf{61.8} & \textbf{74.1} & \multirow{2}{*}{128}  & \multirow{2}{*}{3.97M} & \textbf{60.2} & \textbf{50.8} & \textbf{65.9} & \textbf{78.1} \\
&          &       &   \textcolor{red}{\small{( +0.6)}}    &   \textcolor{red}{\small{( + 0.4)}}    & \textcolor{red}{\small{( +0.3)}} & \textcolor{red}{\small{( +1.1)}} &          &       &   \textcolor{red}{\small{( +0.2)}}    &   \textcolor{red}{\small{( +0.8)}}    & \textcolor{red}{\small{( +0.9)}} & \textcolor{red}{\small{( +0.1)}} \\ \hline
\multirow{2}{*}{TeLM}       & \multirow{2}{*}{121}  & \multirow{2}{*}{3.85M} & \textbf{54.5} & \textbf{45.5} & \textbf{59.5} & \textbf{71.7} & \multirow{2}{*}{121}  & \multirow{2}{*}{7.26M} & \textbf{59.0} & \textbf{50.6} & \textbf{63.8} & \textbf{74.7} \\
&          &       &   \textcolor{red}{\small{( +3.9)}}    &   \textcolor{red}{\small{( +3.4)}}    & \textcolor{red}{\small{( +4.5)}} & \textcolor{red}{\small{( +3.6)}} &          &       &   \textcolor{red}{\small{( +2.2)}}    &   \textcolor{red}{\small{( +1.9)}}    & \textcolor{red}{\small{( +2.7)}} & \textcolor{red}{\small{( +2.7)}} \\ \hline
\multirow{2}{*}{TeAST}      & \multirow{2}{*}{128}  & \multirow{2}{*}{2.13M} & \textbf{56.1} & \textbf{47.3} & \textbf{61.4} & \textbf{72.3} & \multirow{2}{*}{128}  & \multirow{2}{*}{4.87M} & \textbf{59.2} & \textbf{50.5} & \textbf{64.8} & \textbf{76.9} \\ 
&          &       &   \textcolor{red}{\small{( +2.7)}}    &   \textcolor{red}{\small{( +3.4)}}    & \textcolor{red}{\small{( +2.6)}} & \textcolor{red}{\small{( +1.4)}} &          &       &   \textcolor{red}{\small{(10.4)}}    &   \textcolor{red}{\small{( +12.1)}}    & \textcolor{red}{\small{( +10.1)}} & \textcolor{red}{\small{(+8.6)}} \\  \specialrule{1.2pt}{0pt}{0pt}
\end{tabular}}
  \caption{Link prediction results on ICEWS14 and ICEWS05-15. The results of $\heartsuit$ are taken from \citet{2019-tnt}. Other results are obtained from our experiments. ${log}(f(\cdot))$ indicates that our proposed method.}
  \label{tab:result}
\end{table*}

In our experiments, we validate the effectiveness of our proposed method for dealing with TKGs heterogeneity in tensor decomposition on ICEWS14 and ICEWS05-15 datasets. The improvements are marked in red in Table~\ref{tab:result}, highlighting the advancements over the baselines. When our method is applied to different tensor decomposition based TKGE models, they all achieve meaningful improvements in different metrics. This significant improvement confirms our Proposition 1, in which homogeneous factor tensors can be effectively approximated $\mathcal{Y}$ with a low rank. Additionally, the ICEWS05-15 dataset validates the effectiveness of our method in mitigating data heterogeneity, with TeAST notably exhibiting an average improvement of 10.3 points.

In conclusion, the experimental results provide robust evidence supporting Proposition 1. The experimental outcomes validate the theoretical framework of our study and demonstrate that our novel method effectively alleviates data heterogeneity in tensor decomposition and enhances the link prediction performance of these models. More experiments on the large TKG dataset GDELT can be found in \textbf{Appendix~\ref{app-c}}.

\subsection{Quantitative Analysis on Heterogeneity}

In this section, we perform a quantitative analysis to prove the effectiveness of our proposed method. For any factor tensors $\bm{e}_x$ and  $\bm{e}_y$, the skew-symmetric matrices in \(\mathfrak{so}(n)\) are given by
\begin{equation}
    A_x = {log}(f(\bm{e}_x)), \quad A_y = {log}(f(\bm{e}_y)).
\end{equation}

We define the difference between $ \bm{e}_x$  and $\bm{e}_y$  in $\mathfrak{so}(n)$  to be denoted as $d(A_x, A_y)$.
The relationship between the set of skew-symmetric matrices \( \{A_x\} \) obtained from the mapping of a set of vectors \(  \bm{e}_x \) can be described using the operations in the Lie algebra, such as computing their Lie brackets. This process of mapping vectors to $SO(n)$ and then to \(\mathfrak{so}(n)\) transforms them into elements with a unified algebraic structure, mitigating the differences in structural distribution between them.

To quantify the structural differences between the skew-symmetric matrices \( A_x \) and \( A_y \) in \(\mathfrak{so}(n)\), we consider the Frobenius norm of their difference
\begin{equation}
\scalebox{0.86}{$
    \| A_x - A_y \|_F = \sqrt{\text{trace}((A_x - A_y)^T (A_x - A_y))}. 
$}
\end{equation}
This norm provides a measure of the difference between the corresponding  matrices.

Based on the above quantitative formulas, we evaluate on the representative models TComplEx and  TNTComplEx. Specifically,  we calculate the difference between entity and relation, entity and timestamp, and relation and timestamp. As shown in Table~\ref{tab:qa}, we calculate their average distance difference on ICEWS14.

\linespread{1.2}
\begin{table}[h]
\centering
\scalebox{0.8}{
\begin{tabular}{cccc}
\specialrule{1.2pt}{0pt}{0pt}
           & $d(|\mathcal{E}|, |\mathcal{R}|)$ & $d(|\mathcal{E}|, |\mathcal{T}|)$ & $d(\mathcal{R}, |\mathcal{T}|)$ \\ \hline

TComplEx &  15.71 &  7.61 & 15.72  \\ 
TComplEx & \multirow{2}{*}{\textbf{13.74}}  & \multirow{2}{*}{\textbf{6.89}}  & \multirow{2}{*}{\textbf{12.43}} \\ 
$+{log}(f(\cdot))$ &  &  &   \\  \hline 
TNTComplEx &  22.20 &  5.82 & 22.06  \\ 
TNTComplEx & \multirow{2}{*}{\textbf{17.92}}  & \multirow{2}{*}{\textbf{5.61}}  & \multirow{2}{*}{\textbf{17.10}} \\ 
$+{log}(f(\cdot))$ &  &  &   \\ \specialrule{1.2pt}{0pt}{0pt}
\end{tabular}}
    \caption{Quantitative analysis results for TComplEx \textit{vs.} TComplEx$+{log}(f(\cdot))$ and  TNTComplEx \textit{vs.} TNTComplEx$+{log}(f(\cdot))$.}
    \label{tab:qa}
\end{table}

As shown in Table~\ref{tab:qa}, for the standard TComplEx model, the average difference in distance between entities and relations is 15.71, between entities and timestamps is 7.61, and between relations and timestamps is 15.72. These results demonstrate significant differences in quantification between different types of  embeddings.

Further, when our method ${log}(f(\cdot))$ is employed in the TComplEx model, we observe a reduction in all three distance difference. This significant improvement points to the effectiveness of our method in mitigating the heterogeneity among different factor tensors. Similarly, the results of the TNTComplEx model support the above statement.

\subsection{Visualisation  Analysis}

\begin{figure}[h]
    \centering
    \subfigure[TComplEx]{
      \includegraphics[width=0.22\textwidth]{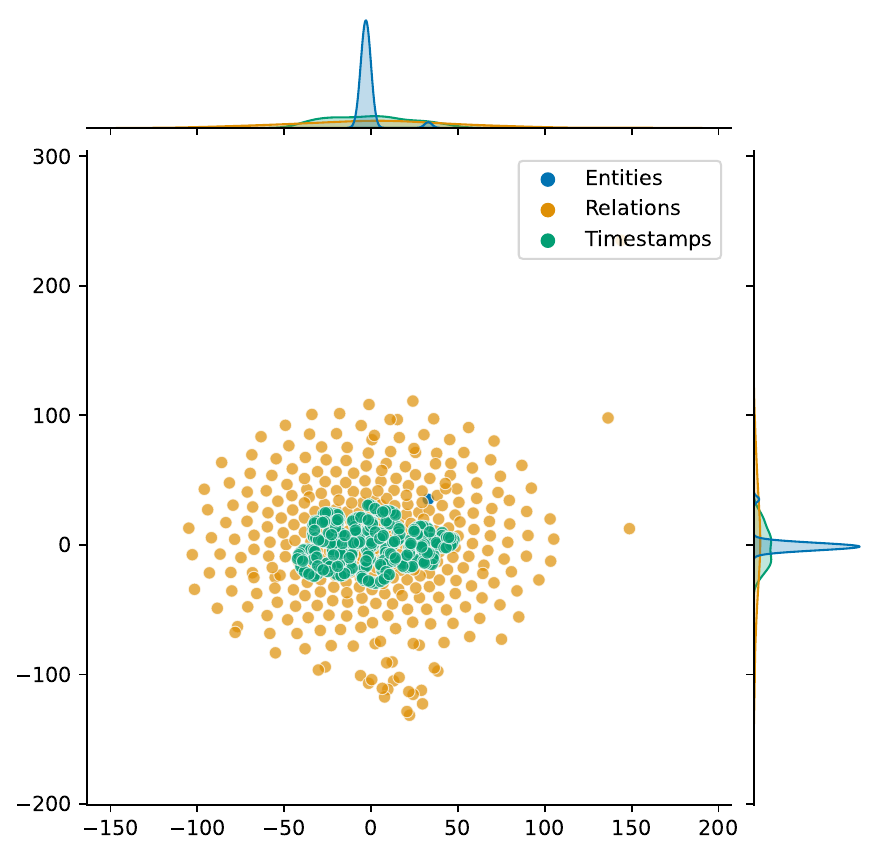}
    }
    \subfigure[TComplEx+$\log(f(\cdot))$]{
      \includegraphics[width=0.22\textwidth]{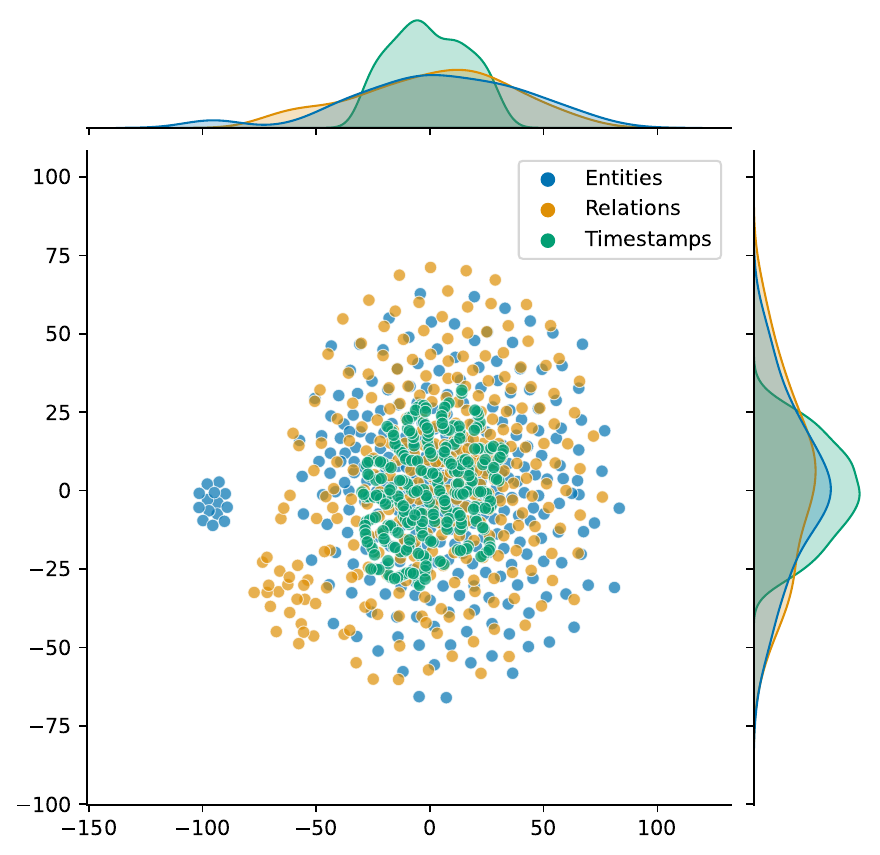} 
    }
    \subfigure[TNTComplEx]{
      \includegraphics[width=0.22\textwidth]{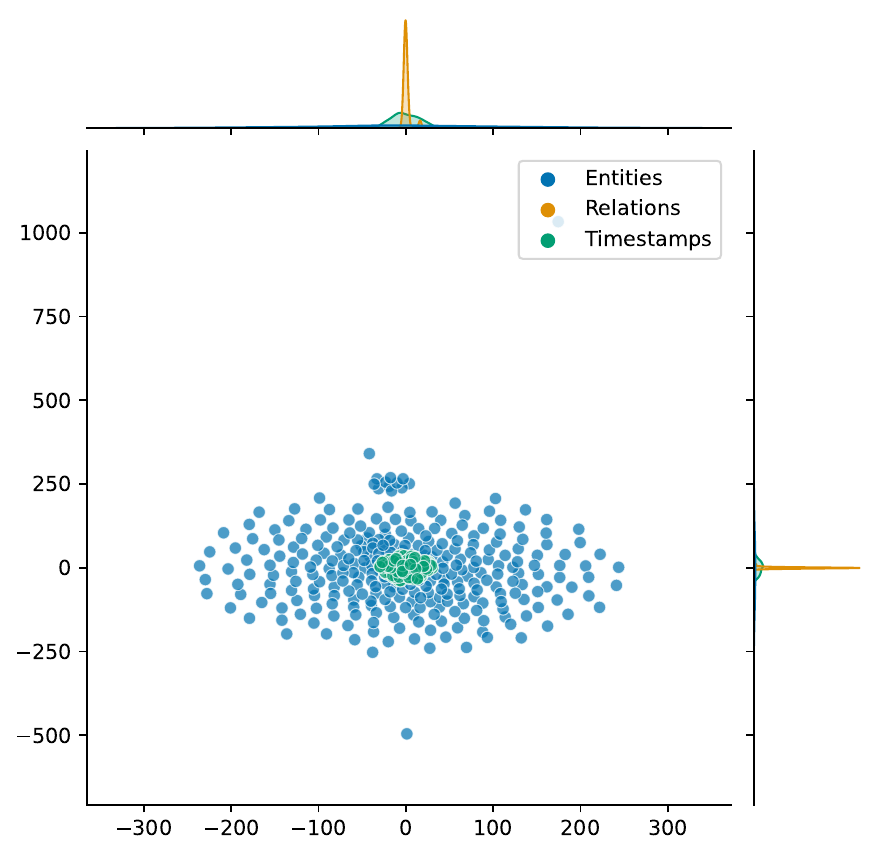}
    }
    \subfigure[TNTComplEx+$\log(f(\cdot))$]{
      \includegraphics[width=0.22\textwidth]{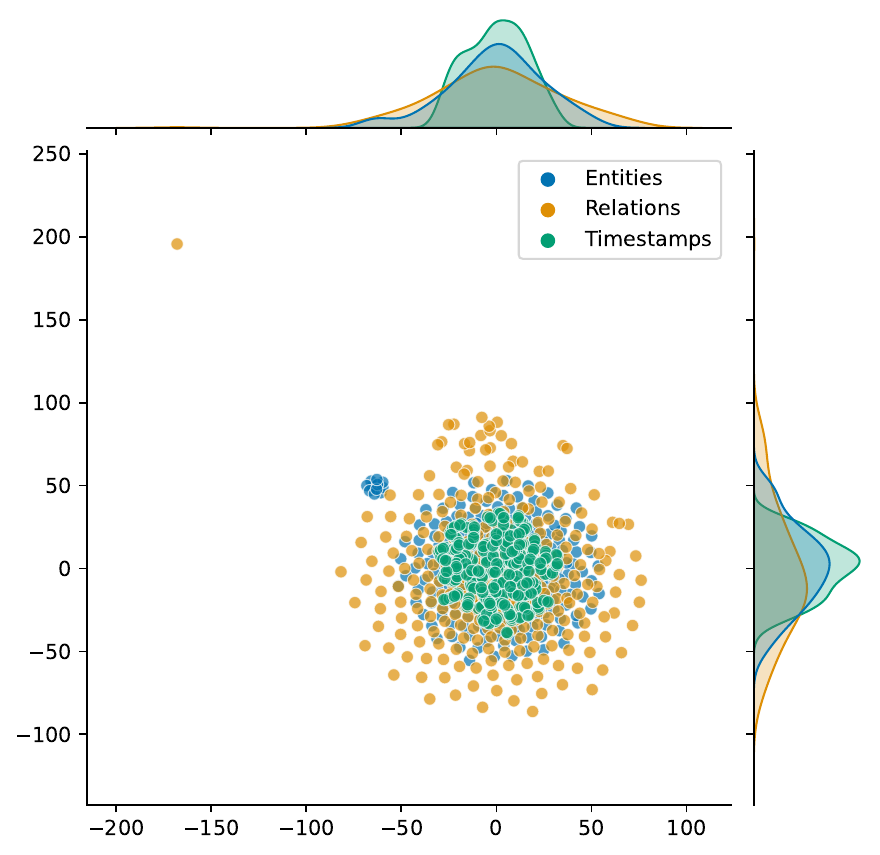} 
    }
    \caption{Visualisations of the learned entity, relation and timestamp embeddings on ICEWS14.}
    \label{fig:tsne}
  \end{figure}

To further verify that our method can effectively mitigate the heterogeneity between factor embeddings, we utilize t-SNE~\cite{van2008visualizing} to visualize the learned entity, relation, and timestamp embeddings. As shown in Figure~\ref{fig:tsne}, we can observe that the learned factor embeddings through our method exhibit a trend towards homogeneity. This further demonstrates the inherent heterogeneity present among different types of embeddings in TKGE based on tensor decomposition. Our method effectively mitigates this issue, demonstrating that meaningful performance improvements can be achieved.

\subsection{Effect of Rank}

\begin{figure}[!htb]
    \centering
    \subfigure[MRR]{\label{fig:grad-a}
      \includegraphics[width=0.22\textwidth]{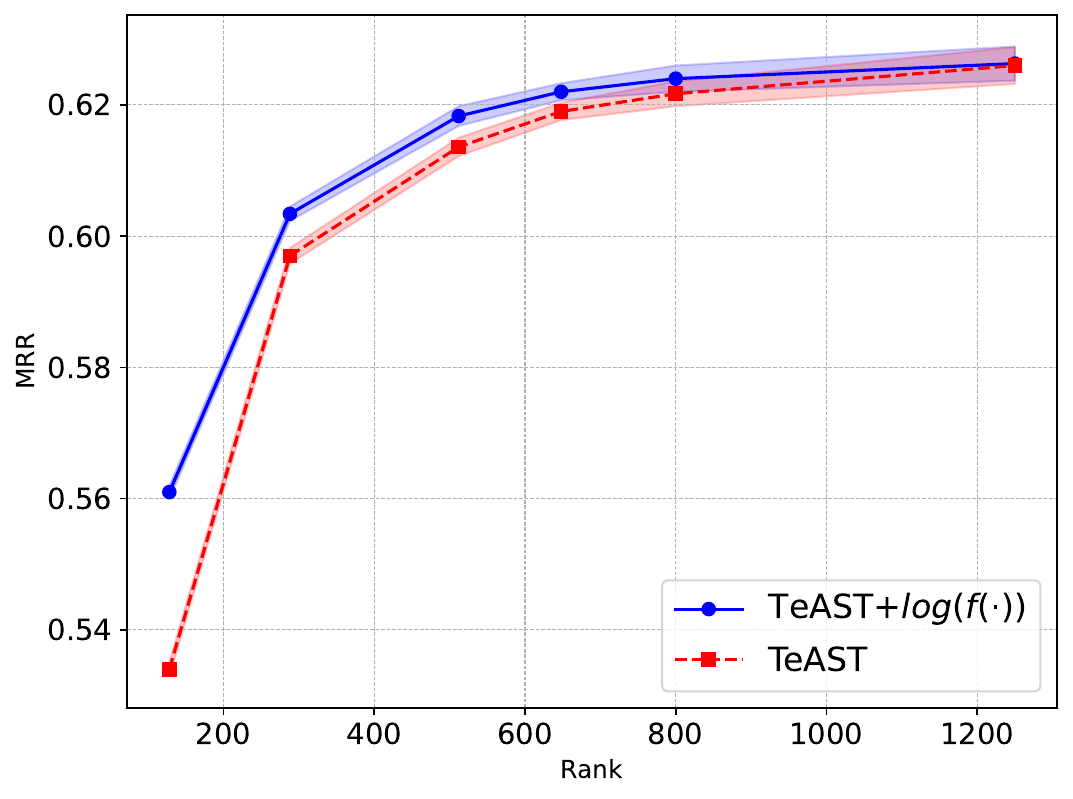}
    }
    \subfigure[Hits@1]{\label{fig:grad-b}
      \includegraphics[width=0.22\textwidth]{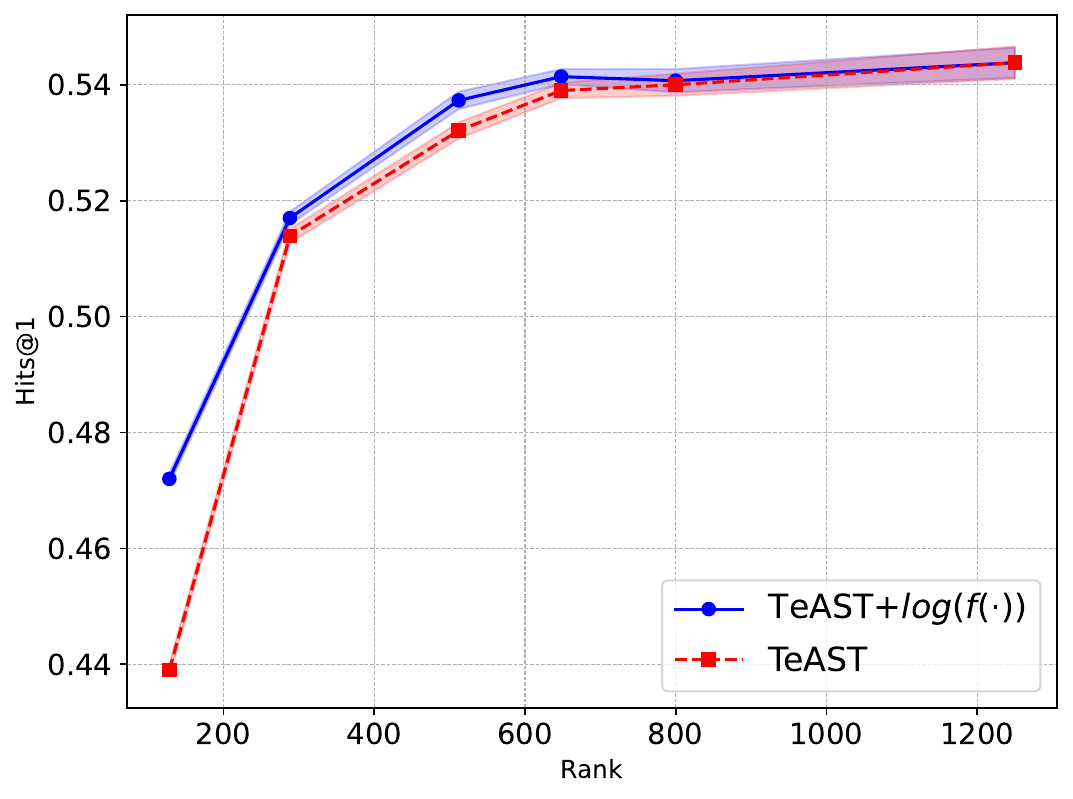} 
    }
    \subfigure[Hits@3]{\label{fig:grad-c}
      \includegraphics[width=0.22\textwidth]{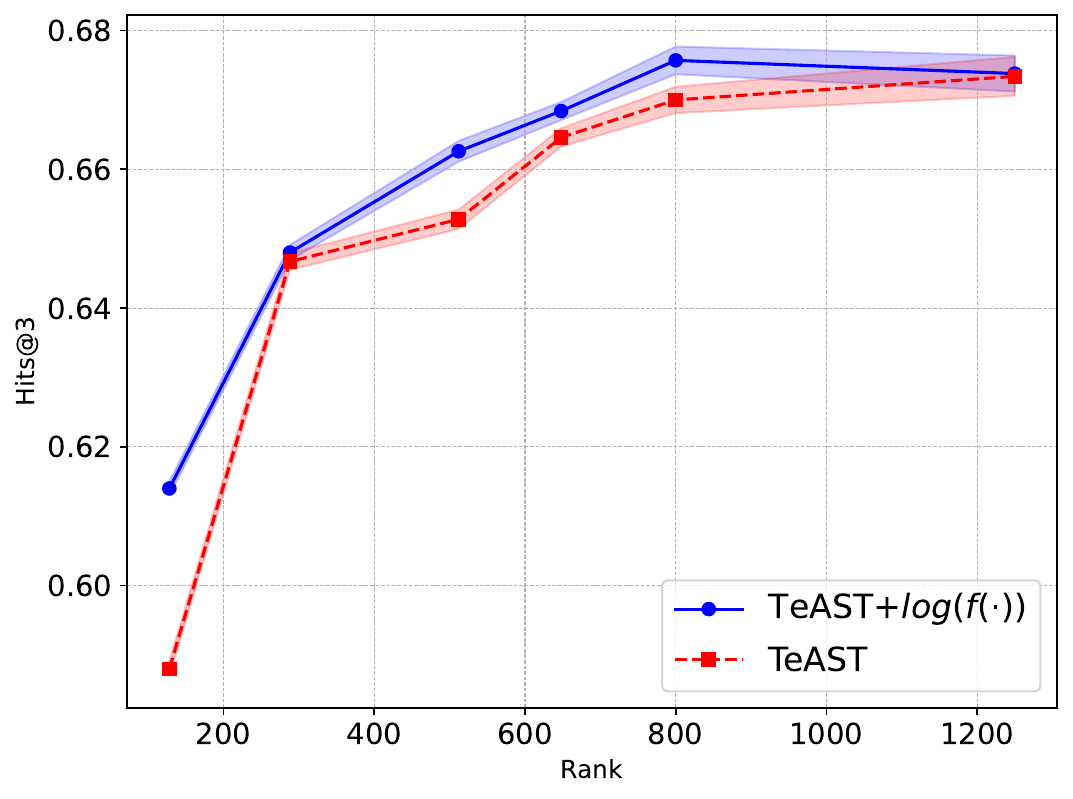} 
    }
    \subfigure[Hits@10]{\label{fig:grad-d}
      \includegraphics[width=0.22\textwidth]{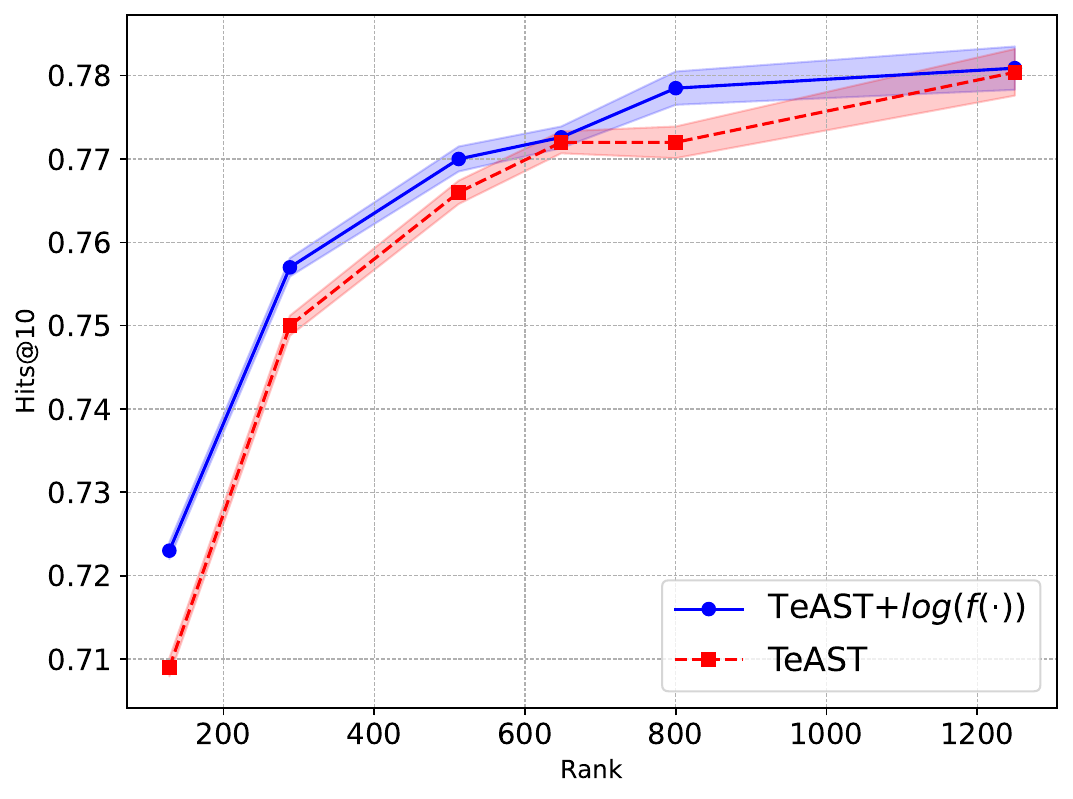} 
    }
    \caption{Results of  TeAST and TeAST$+{log}(f(\cdot))$ with different rank on  ICEWS14.}
    \label{fig:grad}
  \end{figure}

In this work, we compare the performance of the standard TeAST~\cite{li-etal-2023-teast} model with our proposed TeAST model enhanced by \( +\log(f(\cdot)) \) across different rank values on ICEWS14. As shown in Figure~\ref{fig:grad},  we observe that the performance of both models improves with the increase of rank values. However, after the rank value reaches 800, the pace of performance improvement slows down. This is because the representation capacity of the model reaches saturation at a certain level, beyond which the marginal benefits of increasing rank diminish. Additionally, higher rank values might lead to overfitting, especially in cases of sparse data, negatively affecting the model's generalization ability on unseen data.

\section{Conclusion}
In this study, we are the first to introduce methods to mitigate heterogeneity in factor tensors within tensor decomposition-based TKGE models. We reveal that the heterogeneity primarily stems from diverse semantic content among elements (entities, relation and timestamps), which impedes the effective fusion of factor tensors and limits link prediction accuracy. We prove that homogeneous tensors are more effective than heterogeneous tensors in tensor fusion and approximating the target for tensor decomposition based TKGE methods. Our method maps factor tensors onto a smooth Lie group manifold to standardize their distribution and mitigate heterogeneity without increasing model complexity. Our experimental results demonstrate the effectiveness of this method in mitigating tensor heterogeneity and enhancing performance.  We hope that this work can offer fresh insights for research in the field of TKGE.

\section*{Limitations}

In this paper, we investigate the effect of the heterogeneity among factor tensors on link prediction in tensor decomposition based TKGE models. We mainly focus on addressing the issue of heterogeneity among elements within TKGs, which is recognized as a key challenge in this domain. Moreover, similar to the majority of TKGE models, our method is unable to process new entities that are not present in the training data.

\section*{Acknowledgments}
This work was funded by National Natural Science Foundation of China (Grant No. 62366036), National Education Science Planning Project (Grant No. BIX230343), Key R\&D and Achievement Transformation Program of Inner Mongolia Autonomous Region (Grant No. 2022YFHH0077), The Central Government Fund for Promoting Local Scientific and Technological Development (Grant No. 2022ZY0198), Program for Young Talents of Science and Technology in Universities of Inner Mongolia Autonomous Region (Grant No. NJYT24033), Inner Mongolia Autonomous Region Science and Technology Planning Project (Grant No. 2023YFSH0017), Science and Technology Program of the Joint Fund of Scientific Research for the Public Hospitals of Inner Mongolia Academy of Medical Sciences (Grant No.2023GLLH0035).

\bibliography{anthology}

\appendix

\section{Definition and Discussion of Heterogeneity in TKG}\label{sec:ddh}
In KGs, `heterogeneity' refers to the semantic difference of the entity and relation. Similarly, in TKGs, `heterogeneity' refers to the semantic difference of the entity, relation and timestamp.  The heterogeneity among entities, relations, and time is as follows: (1) The heterogeneity between entities and relations is reflected  in their structural roles within the graph, with entities existing as nodes and relations represented as edges between nodes. (2) The heterogeneity between entities and time is reflected in the fact that entities represent the static components of the graph, while time affects the changes in entity attributes and their relations.  (3) The heterogeneity between relations and time is reflected in that relations delineate the interactions among entities, while time characterizes the temporal aspects of these interactions, specifying when they occur and their duration. Recent work~\cite{zhang2019heterogeneous,li2021learning,cai2022multi} also indicates that KGs have an intrinsic property of heterogeneity, which contains various types of entities and relations. Since TKG extends the KG paradigm, they inherently exhibit this heterogeneity as well. Additionally, TKGs incorporate temporal information, which further contributes to time heterogeneity.

The heterogeneity in TKG leads to the learned factor tensor expliciting different distributions in tensor decomposition based TKGE methods. Unlike previous works~\cite{wu2020temp,park2022evokg}, we do not propose a model for modeling heterogeneous TKGs, but rather a unified approach for mitigating heterogeneity among  entities, relations and timestamps via Lie group.

\section{Statistics of Datasets}

Statistics of all the datasets used in this work are listed in Table~\ref{tab:dataset}. $\mathcal{E}$ denotes the set of entities, $\mathcal{R}$ denotes the set of relations, and $\mathcal{T}$ denotes the set of timestamps.

\label{sec:appendix}
\linespread{1.2}
\begin{table}[h]
	\centering
	\scalebox{0.8}{

	\begin{tabular}{c|ccc} \specialrule{1.2pt}{0pt}{0pt}
				& \textbf{ICEWS14} & \textbf{ICEWS05-15}   &\textbf{GDELT}   \\ \hline
	$\mathcal{E}$         & 7,128   & 10,488     &  500     \\
	$\mathcal{R}$         & 230     & 251      &20        \\
	$\mathcal{T}$         & 365     & 4017     & 366        \\\hline
	\#Train          & 72,826  & 386,962  &  2,735,685  \\
	\#Vaild          & 8,963   & 46,092   & 341,961    \\
	\#Test          & 8,941   & 46,275    &  341,961  \\ \specialrule{1.2pt}{0pt}{0pt}
	\end{tabular}}
	\caption{Statistics of ICEWS14, ICEWS05-15 and GDELT datasets in the experiment.}
	\label{tab:dataset}
\end{table}

The ICEWS14 and ICEWS05-15 datasets exhibit heterogeneity across multiple dimensions, encompassing a wide array of entities, relations, and temporal variations. These datasets include diverse entities such as countries, governmental bodies, individuals, and organizations, each with unique attributes and patterns of behavior. The relations captured within these datasets are equally varied, detailing interactions ranging from diplomatic engagements to military conflicts, each bearing distinct characteristics and impacts. Furthermore, the chronological recording of events introduces a dynamic aspect to the data, with entities and their interrelations evolving over different times.

\section{Results on Larger Dataset}\label{app-c}

The above experiments have validated that our proposed method can improve the TKGE performance on the high-heterogeneity KGs by mitigating heterogeneity among factor tensors for tensor decomposition based methods. To further validate the effectiveness, we conduct experiments on a larger and more challenging TKG dataset GDELT. The GDELT covers only 500 most common entities and 20 most frequent relations, while the number of quadruples achieves 2M. This is reflected in the denser relations between entities in KGs. Hence, the GDELT  dataset is a challenging large-scale TKG. 

 We chose TComplEx  and TNTComplEx models as the backbone model in the experiment on GDELT. The results are shown in Table~\ref{tab:result3}. From Table~\ref{tab:result3}, we observe that there is significant performance improvement in terms of H@1, H@3, and H@10. This proves that the proposed method can effectively diminish the heterogeneity among the factor tensors in TKGE. It exemplifies the potential of our method in handling large-scale TKGs.

\begin{table}[h]
\centering
\scalebox{0.6}{
\begin{tabular}{ccccccc}
\specialrule{1.2pt}{0pt}{0pt}
           & \multicolumn{6}{c}{\textbf{GDELT}}         \\ \cline{2-7} 
           & \emph{rank} & \emph{Para.} & MRR   & H@1   & H@3   & H@10  \\ \hline
TComplEx    & 128 & 0.23M  & 21.3  & 13.4  & 22.7  & 36.5  \\
TComplEx +${log}(f(\cdot))$   & 128 & 0.23M  & \textbf{22.7} & \textbf{14.7} & \textbf{24.3} & \textbf{38.3} \\ \hline
TNTComplEx   & 128 & 0.24M  & 21.9  & 13.9  & 23.3  & 37.4  \\ 
TNTComplEx $+{log}(f(\cdot))$ & 128 & 0.24M  & \textbf{22.1} & \textbf{14.0} & \textbf{23.5} & \textbf{37.6} \\  \specialrule{1.2pt}{0pt}{0pt}
\end{tabular}}
 \caption{Link prediction results on GDELT.}
  \label{tab:result3}
\end{table}

\section{Method Efficiency Comparison}

Since we implement logarithmic mapping in our implementation of Lie group mapping using the following method, our method is theoretically a linear operation with $O(n)$ time complexity. The following table shows the training time required for our method on the ICEWS05-15 dataset compared to other methods, confirming the linear operation of our method without significantly increasing the computational cost. As shown in Table~\ref{table:training_comparison}, we can observe that the training time for models using the ${log}(f(\cdot))$ method shows a slight increase. However, considering the potential performance improvements brought by our method, this additional time cost is acceptable.

\begin{table}[t]
\centering
\scalebox{0.8}{
\begin{tabular}{lcc}
\hline
\textbf{Method} & \textbf{\textit{Para.}} & \textbf{Train-time} \\
\hline
TComplEx & 3.84M & 30 min \\
TComplEx +${log}(f(\cdot))$ & 3.84M & 32 min \\
TNTComplEx & 3.97M & 32 min \\
TNTComplEx +${log}(f(\cdot))$ & 3.97M & 33 min \\
TeLM & 7.26M & 35 min \\
TeLM +${log}(f(\cdot))$ & 7.26M & 36 min \\
TeAST & 4.87M & 34 min \\
TeAST +${log}(f(\cdot))$ & 4.87M & 35 min \\
\hline
\end{tabular}}
\caption{Comparison of training times and parameters for different TKGE models on ICEWS05-15.}
\label{table:training_comparison}
\end{table}

\end{document}